\documentclass{article}

\PassOptionsToPackage{numbers, compress}{natbib}
 \usepackage[dblblindworkshop, final]{neurips_2025}
 \workshoptitle{Non-Euclidean Foundation Models and Geometric Learning}

\usepackage[utf8]{inputenc} 
\usepackage[T1]{fontenc}    
\usepackage{hyperref}       
\usepackage{url}            
\usepackage{booktabs}       
\usepackage{amsfonts}       
\usepackage{nicefrac}       
\usepackage{microtype}      
\usepackage{xcolor}         
\usepackage{algorithm}
\usepackage{amsmath}
\usepackage{amsthm}
\usepackage{algpseudocode}

\usepackage{booktabs}
\usepackage{colortbl}
\usepackage{graphicx}
\usepackage{float}
\usepackage{mathtools}

\newtheorem{theorem}{Theorem}

\newtheorem{corollary}{Corollary}
\newtheorem{definition}{Definition}

\newtheorem*{repthm}{Theorem~\ref{thm:entropy-generalization}}
\newtheorem*{repthm2}{Theorem \ref{thm:iblm}}
\newtheorem*{repcor}{Corrolary \ref{cor:discrete-entropy-lower}}

\title{Memorization-Compression Cycles Improve Generalization}

\author{%
  Fangyuan Yu  \\
  Thoughtworks \\
  \texttt{fangyuan.yu@thoughtworks.com}
}

\begin{document}

\maketitle

\begin{abstract}
We prove theoretically that generalization improves not only through data scaling but also by compressing internal representations. To operationalize this insight, we introduce the Information Bottleneck Language Modeling (IBLM) objective, which reframes language modeling as a constrained optimization problem: minimizing representation entropy subject to optimal prediction performance. Empirically, we observe an emergent memorization–compression cycle during LLM pretraining, evidenced by oscillating positive/negative gradient alignment between cross-entropy and Matrix-Based Entropy (MBE), a measure for representation entropy.  This pattern closely mirrors the predictive–compressive trade-off prescribed by IBLM and also parallels the biological alternation between active learning and sleep consolidation. Motivated by this observation, we propose Gated Phase Transition (GAPT), a training algorithm that adaptively switches between memorization and compression phases. When applied to GPT-2 pretraining on FineWeb dataset, GAPT reduces MBE by $50\%$ and improves cross-entropy by $4.8\%$. GAPT improves OOD generalization by 35\% in a pretraining task on arithmetic multiplication. In a setting designed to simulate catastrophic forgetting, GAPT reduces interference by compressing and separating representations, achieving a 97\% improvement in separation — paralleling the functional role of sleep consolidation in biological learning process.
\end{abstract}

\section{Introduction}

Generalization occurs when learning from one task improves performance on another. The pursuit of generalization in pre-training LLM \cite{radford2019gpt2} has historically focused on scaling up data and parameter size \cite{kaplan2020scalinglawsneurallanguage}, in post-training, Reinforcement Learning with Verifiable Reward (RLVR) \cite{deepseekai2025deepseekr1} has gained attention. However, high quality data has been exhausted after years of tireless scraping, and RLVR is shown to only trims knowledge from baseline model \cite{yue2025doesreinforcementlearningreally} instead of 'incentivize new reasoning pattern'. 

We establish a theoretical upper bound on generalization error for deep learning models, indicating representation entropy as another dimension for improving generalization besides scaling data.
\begin{theorem}\textbf{Upper Bound on Generalization Error.}
\label{thm:entropy-generalization}
Let $X \in \mathcal{X}$ and $Y \in \mathcal{Y}$ be random variables with an unknown joint distribution $P(X, Y)$, and suppose $X$ is discrete with finite cardinality. Let $f$ be a neural network with $L$ intermediate representations forming a Markov chain:
\[
X \rightarrow R_1 \rightarrow \cdots \rightarrow R_L \rightarrow \hat{Y}
\]
Where $R_{l}$ is internal representations, $\hat{Y}$ is the prediction of the network. Then, for any dataset $\mathcal{D}_N = \{(x_i, y_i)\}_{i=1}^N$ sampled i.i.d. from $P(X,Y)$, the generalization error satisfies for $\alpha \in [1,+\infty)$:
\begin{equation}
\underbrace{\mathcal{L}_{P(X,Y)}(f, \ell)}_{\mathclap{\text{Generalization Error}}}
\leq 
\underbrace{\hat{\mathcal{L}}_{P(X,Y)}(f, \ell, \mathcal{D}_N)}_{\mathclap{\text{Empirical Error}}}
+ 
\underbrace{\mathcal{O}\left( \log N \cdot \min_{1 \leq l \leq L} 2^{\alpha \cdot H(R_l)} \, / \, \sqrt{N} \right)}_{\mathclap{\text{Upper Bound on Generalization Gap}}}
\end{equation}
\end{theorem}
Intuitively, generalization performance of neural network can be improved by either increasing training data size, or by reducing entropy of its internal representations. We refer minimization of $H(R_{l})$ as compression, and minimizing empirical loss as memorization.

The Information Bottleneck (IB) framework \cite{tishby2015deeplearninginformationbottleneck} characterizes the optimal representation as one that discards as much information from the input as possible, while preserving all information relevant to the output. Motivated by this, we introduce the Information Bottleneck Language Modeling (IBLM) objective, along with a theorem showing equivalence of IBLM with IB framework under language modeling. 

\begin{definition}\textbf{Information Bottleneck Language Modeling (IBLM)}. 
Given a language model with internal representations $R_{1:L}$ and output token variable $Y$, the IBLM objective is:
\begin{equation}
\begin{aligned}
\min \quad & \sum_{l=1}^{L} H(R_l) \\
\text{s.t.} \quad & \hat{Y} = \arg\min_{\hat{Y}} H(Y | \hat{Y})
\end{aligned}
\label{eq:iblm}
\end{equation}
\end{definition}
\begin{theorem}
\label{thm:iblm}
The IBLM objective defined in Equation~\ref{eq:iblm} is equivalent to the classical Information Bottleneck formulation under the language modeling setting.
\end{theorem}

Note that $H(Y|\hat{Y})$ is the cross-entropy (CE) loss in conventional language modeling task \cite{radford2019gpt2}. In short, LBLM requires maximal compressing internal representation under optimal cross-entropy.  To explicitly calculate $H(R)$, we adopt Matrix-based entropy (MBE), first proposed in \cite{giraldo2014entropy}, given a matrix $R \in \mathbb{R}^{s \times d}$ concatenating $s$ token representations, MBE is given by
\[
S_{\alpha}(T) = \frac{1}{1-\alpha} \log \Big(\sum_{i}(\frac{\lambda_{i}(K)}{\text{Tr}(K)})^{\alpha}\Big)
\]
where $K = R R^{T}$ is the Gram matrix. MBE essentially provide a continuous measure of matrix rank, by calculating entropy on distribution of singular values. It has been shown to exhibit a strong correlation with embedding quality \cite{skean2025layerlayeruncoveringhidden}.

Previous work \cite{shwartzziv2017openingblackboxdeep} showed that in neural networks, training often follows a two-phase trend: an initial memorization phase with rapid loss reduction, followed by a monotonic decrease in representation entropy $H(R)$, interpreted as compression. This suggests a clean, sequential separation between fitting and compressing.

Our analysis of GPT pretraining reveals a richer dynamic: the cosine similarity between CE and MBE gradients oscillates between positive and negative, indicating a cyclic alternation between memorization (representation expansion) and compression. Rather than progressing through distinct phases, learning continuously revisits these states. Notably, this local oscillation coexists with a global decline in MBE, suggesting that compression accumulates over time even as the model periodically re-engages in memorization. We term this the memorization–compression cycle.

In biological neural systems, learning proceeds cyclically, alternating between active acquisition and sleep-driven consolidation. A seminal study \cite{gonalez2020sleepprotectmemories} showed that sleep mitigates interference between conflicting memories by compressing and reorganizing them—allocating distinct subsets of synaptic weights to each. When two conflicting tasks were presented sequentially, awake learning alone led to catastrophic forgetting, while sleep consolidation enabled the formation of separable representations, outperforming even interleaved training. Inspired by this, we propose Gated Phase Transition (GAPT), a training algorithm that alternates between memorization (minimizing cross-entropy) and compression (minimizing cross-entropy and Matrix-Based Entropy). GAPT dynamically switches phases based on learning signals, approximating the IBLM objective and functionally mirroring biological consolidation to resolve representational conflict.

GAPT delivers consistent improvements across domains. In LLM pretraining on the FineWeb dataset, it reduces Matrix-Based Entropy (MBE) by an average of $70.5\%$ across layers while improving cross-entropy by $4.8\%$, outperforming standard language modeling and closely approximating the IBLM trade-off. In an arithmetic generalization task, GAPT reduces test cross-entropy by $35\%$ and MBE by $47\%$ when trained on 1–3 digit multiplication and evaluated on 4–6 digits, supporting Theorem \textbackslash{}ref\{thm:entropy-generalization\}. Finally, in a synthetic setup with gradient interference, GAPT improves representational separation by 97\% and reduces MBE by $91\%$ relative to a mixed baseline—closely mirroring the conflict resolution behavior observed during biological sleep.

Our work makes several key contributions:
\begin{itemize}
    \item \textbf{Theoretical Foundation}: We derive an upper-bound on generalization error showing that reducing representation entropy can improve generalization, alongside scaling data.
    
    \item \textbf{IBLM Objective}: We formulate Information Bottleneck Language Modeling (IBLM) as a constrained optimization problem, unifying representation entropy and cross entropy targets.
    
    \item \textbf{GAPT Algorithm}: We propose Gated Phase Transition (GAPT), a algorithm to solve IBLM alternates between memorization and compression phases based on dynamic training signals.
    
    \item \textbf{Empirical Results}: We show that GAPT improves LLM pre-training, arithmetic generalization, and conflict resolution.
    
    \item \textbf{Biological Analogy}: We relate the memorization–compression cycle in LLMs to the awake–sleep consolidation cycle in biological systems and validate compression's similarity to consolidation.
\end{itemize}

The remainder of the paper expands on these contributions: Section \ref{section:theory} presents our theoretical framework and generalization bound; Section \ref{section:approach} details the GAPT algorithm; Section \ref{section:experiments} presents empirical results, including the memorization–compression cycle and GAPT's effectiveness; Section \ref{section:relevant_work} discusses relevant work.

\section{Theory}
\label{section:theory}
\begin{corollary}[Entropy Lower Bound for Finite Discrete Random Variables]
Let \( X \) be a discrete random variable with finite support \( \Omega \), where \( |\Omega| = n \), and assume that \( P(X = x) > 0 \) for all \( x \in \Omega \). Then there exists a constant \( \beta \in (0,1] \) such that:
\[
H(X) \geq \beta \cdot \log_{2} |\Omega|.
\]
\label{cor:discrete-entropy-lower}
\end{corollary}
Proof of Corollary \ref{cor:discrete-entropy-lower} can be found in Appendix \ref{section:appendix}. 

\begin{repthm}\textbf{Upper Bound on Generalization Error.}
Let $X \in \mathcal{X}$ and $Y \in \mathcal{Y}$ be random variables with unknown joint distribution $P(X, Y)$, and suppose $X$ is discrete with finite cardinality. Let $f$ be a neural network with $L$ intermediate representations forming a Markov chain:
\[
X \rightarrow R_1 \rightarrow \cdots \rightarrow R_L \rightarrow \hat{Y}
\]
Then, for any dataset $\mathcal{D}_N = \{(x_i, y_i)\}_{i=1}^N$, there exists $\alpha \in [1,+\infty)$ s.t. the generalization error satisfies
\[
\mathcal{L}_{P(X,Y)}(f, \ell) \leq \hat{\mathcal{L}}_{P(X,Y)}(f, \ell, \mathcal{D}_N) + \mathcal{O}\left( \frac{\log N \cdot \min_{1 \leq l \leq L} 2^{\alpha \cdot H(R_l)} }{\sqrt{N}} \right)
\]
\end{repthm}

\begin{proof}
We begin by recalling the standard formulation of the generalization gap:
\[
\text{Gen}_{P(X,Y)}(f, \ell, \mathcal{D}_N) := \mathcal{L}_{P(X,Y)}(f, \ell) - \hat{\mathcal{L}}_{P(X,Y)}(f, \ell, \mathcal{D}_N)
\]
For discrete input variables $X$, it has been shown in~\cite{shamir2010boundongengap} that the generalization gap is upper-bounded by:
\begin{align*}
\mathcal{L}_{P(X,Y)}(f, \ell) - \hat{\mathcal{L}}_{P(X,Y)}(f, \ell, \mathcal{D}_N) &\leq \mathcal{O}\left(\frac{|\mathcal{X}| \log N}{\sqrt{N}}\right) \\
&\leq \mathcal{O}(\frac{2^{\alpha \cdot H(X)} \log N}{\sqrt{N}})
\end{align*}
where $|\mathcal{X}|$ denotes the cardinality of the input space. Second line follows from Corollory \ref{cor:discrete-entropy-lower}, which states that we have $|\mathcal{X}| = \mathcal{O}(2^{\alpha  \cdot H(X)})$ for $\alpha \in [1,+\infty)$. Let $f$ be a neural network with $L$ intermediate representations forming a Markov chain:
\[
X \rightarrow R_1 \rightarrow \cdots \rightarrow R_L \rightarrow \hat{Y}
\]
Fix any intermediate representation $R_l$, and decompose the network into $f = d \circ e$, where $e: \mathcal{X} \rightarrow \mathcal{R}_l$ maps inputs to $R_l = e(X)$, and $d: \mathcal{R}_l \rightarrow \mathcal{Y}$ predicts the output. Applying the generalization bound to the representation $R_l$, we obtain:
\begin{align}
\mathcal{L}_{P(X,Y)}(f, \ell) &= \mathcal{L}_{P(R_l,Y)}(d, \ell_e) \\
&\leq \hat{\mathcal{L}}_{P(R_l,Y)}(d, \ell_e, \mathcal{D}_N) + \mathcal{O}\left( \frac{2^{\alpha  \cdot H(R_l)} \log N}{\sqrt{N}} \right) \\
&\leq \hat{\mathcal{L}}_{P(X,Y)}(f, \ell, \mathcal{D}_N) + \mathcal{O}\left( \frac{2^{\alpha  \cdot H(R_l)} \log N}{\sqrt{N}} \right)
\end{align}
Since the Markov structure ensures that each $R_l$ is a valid bottleneck in the information flow, we can take the tightest such bound across all layers, yielding:
\[
\mathcal{L}_{P(X,Y)}(f, \ell) \leq \hat{\mathcal{L}}_{P(X,Y)}(f, \ell, \mathcal{D}_N) + \mathcal{O}\left( \frac{\log N \cdot \min_{1 \leq l \leq L} 2^{\alpha  \cdot H(R_l)} }{\sqrt{N}} \right)
\]
This concludes the proof. 
\end{proof}

\begin{repthm2}
The Information Bottleneck Language Modeling (IBLM) objective defined in Equation~\ref{eq:iblm} is equivalent to the classical Information Bottleneck formulation under the language modeling setting.
\end{repthm2}
\begin{proof}
The Information Bottleneck (IB) framework \cite{tishby2000information, tishby2015deeplearninginformationbottleneck} defines the optimal representation $R$ as the solution to:
\begin{align}
\min_{p(r|x)} \quad & I(R; X) \\
\text{s.t.} \quad & I(Y; R) = I(Y; X)
\end{align}
That is, $R$ should discard as much input information as possible while preserving all information relevant to predicting $Y$—the minimal sufficient statistic. In large language models (LLMs), the network forms a deterministic Markov chain: $X \rightarrow R \rightarrow \hat{Y}$. Since $R$ is deterministically computed from $X$, we have:
\begin{align}
I(X; R) &= H(R) \quad \text{(as } H(R|X) = 0 \text{)} \\
I(Y; X) &\geq I(Y; R) \geq I(Y; \hat{Y}) = H(Y) - H(Y|\hat{Y})
\end{align}

Where second line follows from data processing inequality \cite{cover2006elements}. Thus,
\begin{align}
I(Y; X) - I(Y; R) &\leq H(Y|\hat{Y}) - H(Y|X)
\end{align}
Since $H(Y|X)$ is fixed, minimizing $H(Y|\hat{Y})$—i.e., cross-entropy—maximizes $I(Y; \hat{Y})$, satisfying the IB predictive constraint. On the compression side, since $I(X; R) = H(R)$, minimizing representation entropy directly reduces the IB compression term. Hence, minimizing cross-entropy aligns with preserving predictive information, while minimizing entropy enforces compression—together forming the IBLM objective.
\end{proof}

\section{Approach}
\label{section:approach}

While applying a Lagrangian objective (CE + $\lambda$·MBE) is a natural approach to solving the constrained optimization in IBLM, we find it often leads to representation collapse: MBE converges to near-zero, but CE worsens as the model loses structure in its internal representations.

Inspired by the alternation between learning and consolidation in biological systems, we divide training into two phases: memorization, where the model minimizes cross-entropy (CE) loss, and compression, where it minimizes a weighted sum of CE and Matrix-Based Entropy (MBE). We propose Gated Phase Transition (GAPT), a training algorithm that dynamically alternates between these phases. GAPT tracks a global minimum CE loss and per-layer MBE histories, and uses patience-based gating to switch phases. Compression is exited early if CE degrades.

GAPT encourages localized compression—reorganizing existing knowledge without interfering with the acquisition of new information—and ensures that entropy reduction occurs only when it does not hinder memorization.

\begin{algorithm}[H]
\caption{Gated Phase Transition (GAPT)}
\label{alg:gapt}
\begin{algorithmic}[1]
\State \textbf{Input:} losses $\mathcal{L}$, thresholds $\delta$, $\tau$, patience $p_m$, $p_c$
\State \textbf{State:} $\phi \in \{1, 2\}$ (1 = mem, 2 = comp), counters $s_m$, $s_c$, $E_{\min}$, $\text{MBE}_{\min}[i]$
\State Extract $\mathcal{L}_{\text{ce}}$, \{MBE$_i$\}; update $\Delta E \gets E_{\min} - \mathcal{L}_{\text{ce}}$, $E_{\min} \gets \min(E_{\min}, \mathcal{L}_{\text{ce}})$

\If{$\phi = 1$}  \Comment{Memorization}
    \State $s_m \gets 0$ if $\Delta E > \delta$ else $s_m +\!= 1$
    \If{$s_m \ge p_m$}
        \State $\phi \gets 2$, $s_c \gets 0$, $E_{\min} \gets \infty$, $\text{MBE}_{\min}[i] \gets \infty$
    \EndIf

\Else \Comment{Compression}
    \If{$\mathcal{L}_{\text{ce}} > E_{\min} \cdot (1 + \tau)$}
        \State $\phi \gets 1$, $s_m \gets 0$
    \Else
        \State $\Delta M \gets \max_i (\text{MBE}_{\min}[i] - \text{MBE}_i)$
        \For{each $i$} \State $\text{MBE}_{\min}[i] \gets \min(\text{MBE}_{\min}[i], \text{MBE}_i)$ \EndFor
        \State $s_c \gets 0$ if $\Delta M > \delta$ else $s_c +\!= 1$
        \If{$s_c \ge p_c$} \State $\phi \gets 1$, $s_m \gets 0$ \EndIf
    \EndIf
\EndIf
\end{algorithmic}
\end{algorithm}

\section{Experiments}
\label{section:experiments}

\subsection{Natural Compression-Memorization cycle}
Our experimental setup follows the Modded-NanoGPT framework \cite{modded_nanogpt_2024},
We remove FP8 matmul (due to hardware incompatibility with Hopper GPUs) and use a simplified block causal attention mask. All experiments are conducted on 8×L40 GPUs, training a 12-layer GPT model on a 0.73B-token FineWeb training set, evaluated on its corresponding validation split. We train on CE loss only. We log cross-entropy (CE) and Matrix-Based Entropy (MBE) gradients at every training step. For each iteration, we record both CE and MBE gradients across multiple batches and compute the average cosine similarity both (1) across batches (CE vs. CE), and (2) between CE and MBE gradients, within and across batches.

\begin{figure}  
  \centering
  \includegraphics[width=1\linewidth]{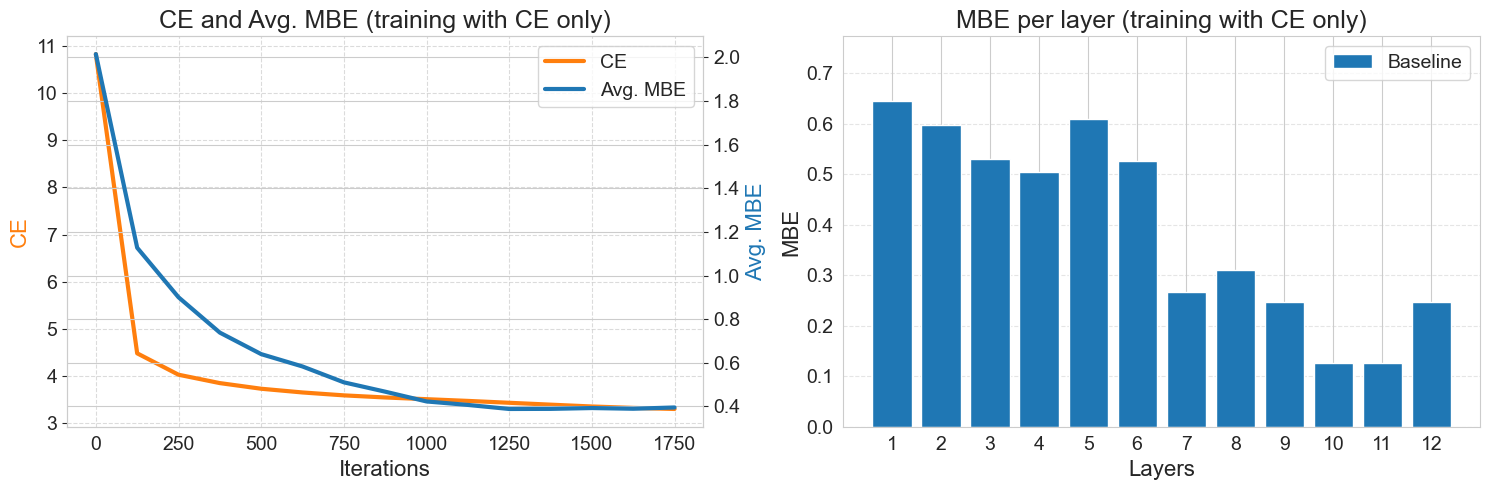}  
  \caption{Left: CE and MBE loss curves during pretraining with CE loss only, showing implicit momentum for representation compression. Right: final per-layer MBE values. Later layers show lower MBE, indicating representation compression.}
  \label{fig:baseline_loss_curve_layer_mbe}
\end{figure}

\begin{figure}  
  \centering
  \includegraphics[width=1\linewidth]{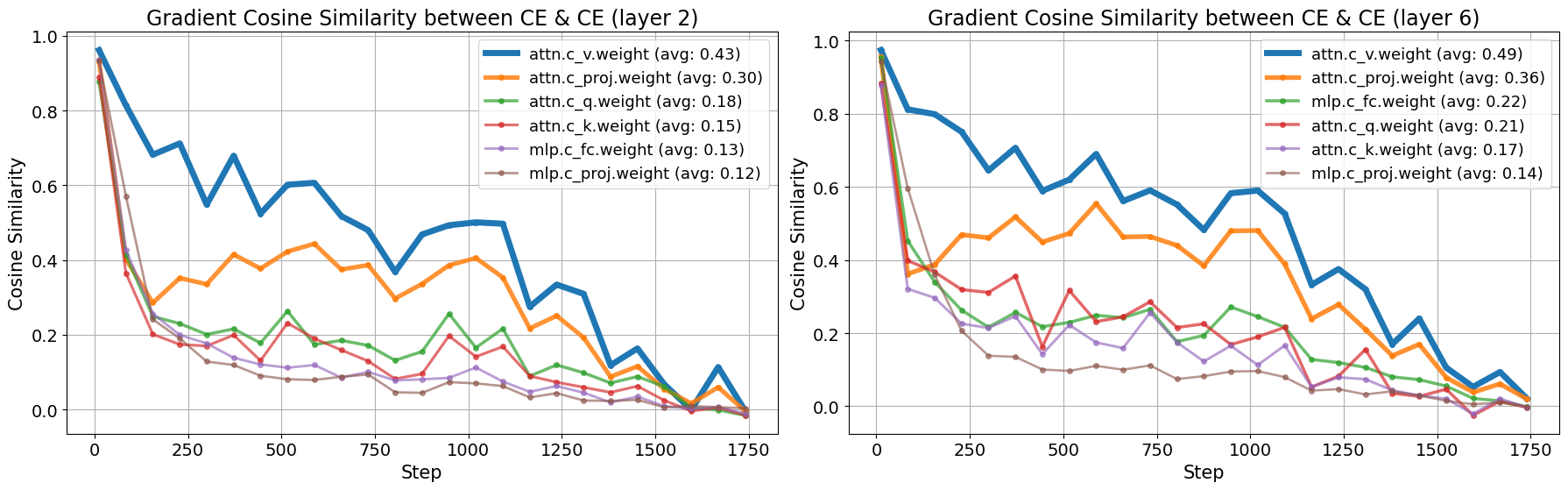}  
  \caption{Cosine similarity between CE gradients across batches. CE gradients become increasingly decorrelated over time, reflecting diminishing shared signal.}
  \label{fig:decline_consistency_ce}
\end{figure}

In Figure \ref{fig:baseline_loss_curve_layer_mbe}, we observe that training with CE loss alone leads to a consistent decrease in MBE across layers, confirming observations from \cite{skean2025layerlayeruncoveringhidden}. We further analyze CE gradient behavior. Figure \ref{fig:decline_consistency_ce}. shows that gradient consistency (measured by cosine similarity across batches) declines over time across all layers. This indicates an increasing signal-to-noise ratio in CE gradients as training progresses.

\begin{figure}[t]  
  \centering
  \includegraphics[width=1\linewidth]{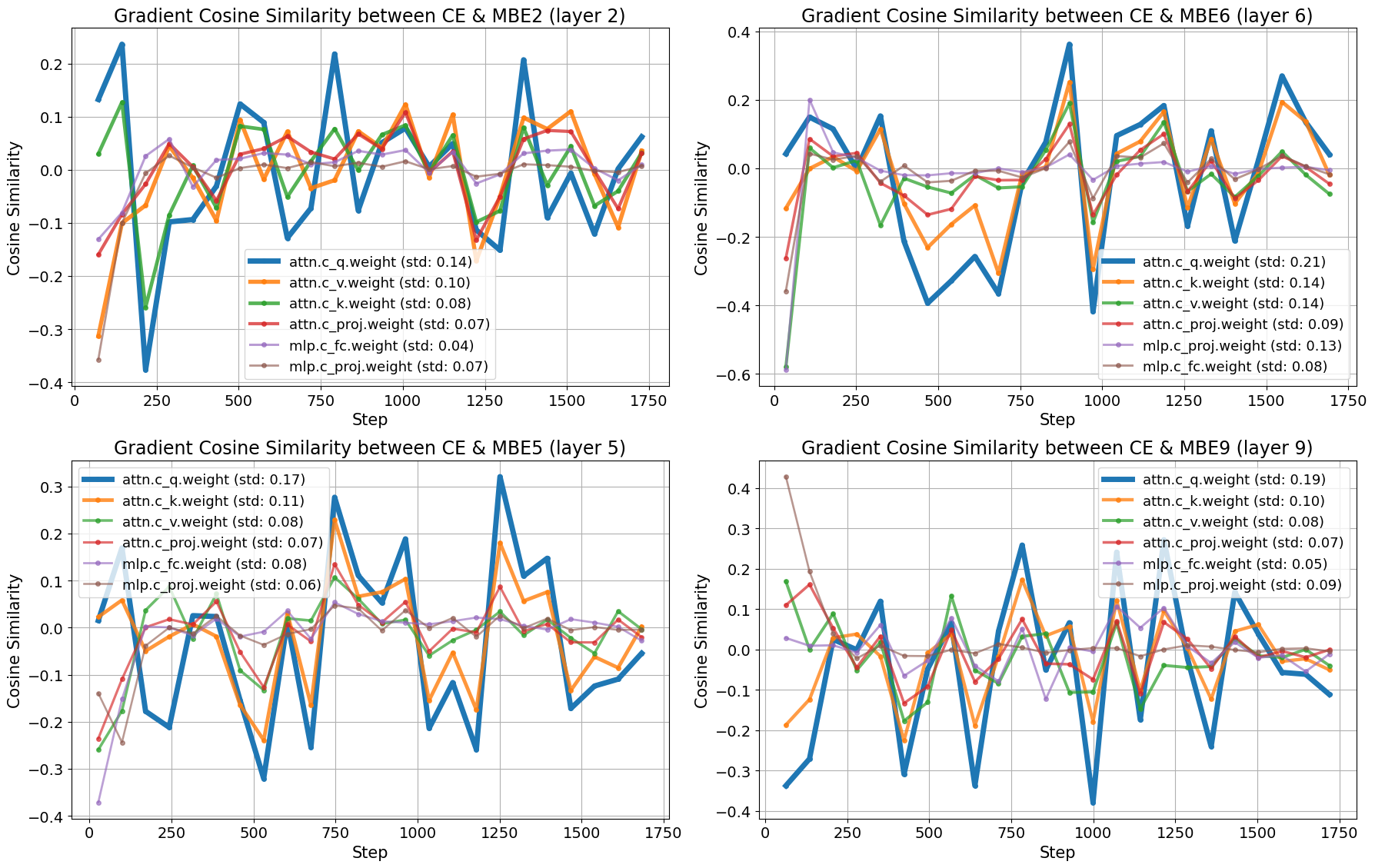}  
  \caption{Cosine similarity between CE and MBE gradients over training. Alternating positive and negative phases indicate emergent memorization–compression cycles.}
  \label{fig:oscillation_entropy_mbe}
\end{figure}

\begin{figure}[t]  
  \centering
  \includegraphics[width=1\linewidth]{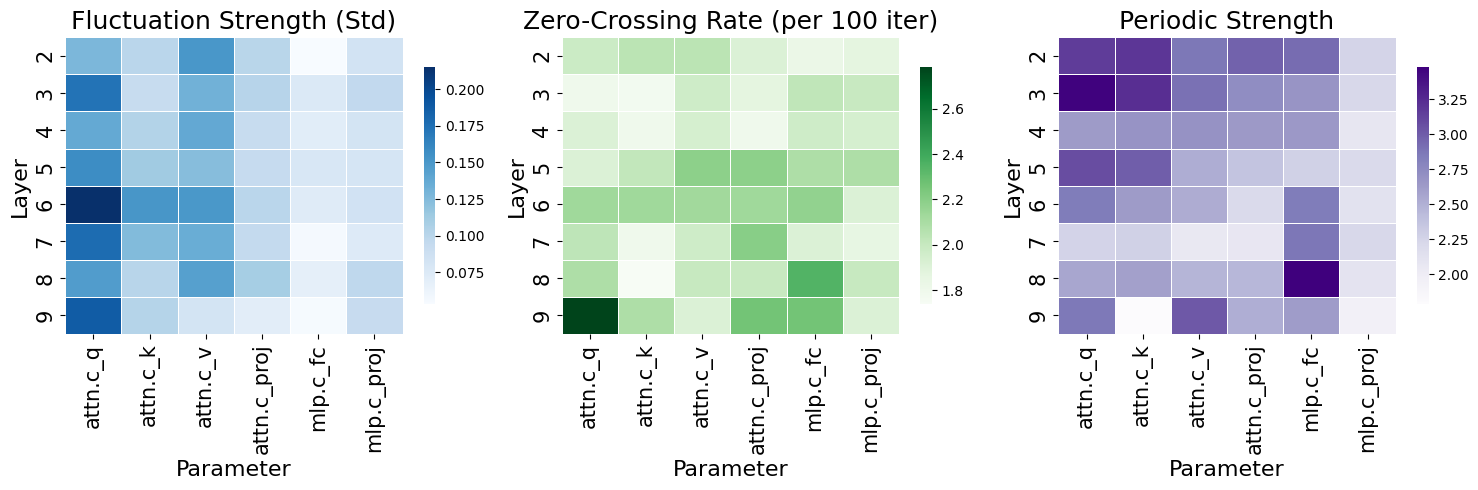}  
  \caption{Oscillation metrics between CE and MBE gradients across layers and parameter groups. Left: standard deviation; center: zero-crossing rate; right: periodic strength (peak-to-mean PSD ratio).}
  \label{fig:oscillation_heatmap}
\end{figure}
We next inspect cosine similarity between CE and MBE gradients. As shown in Figure \ref{fig:oscillation_entropy_mbe}, we observe recurring sign flips that indicate an alternation between memorization and compression phases, even without explicitly optimizing for entropy. To characterize this oscillation, we analyze the gradient signal across training using three measures: standard deviation (oscillation strength), zero-crossing rate (oscillation frequency), and peak-to-average power ratio from the power spectral density (periodicity strength). Figure \ref{fig:oscillation_heatmap} summarizes these metrics across layers and parameters. We find that attention parameters exhibit stronger and more frequent oscillations than MLP parameters. Interestingly, later layers show higher oscillation frequency, while no layer demonstrates strong rhythmic periodicity—suggesting that oscillation is irregular and state-driven, rather than strictly periodic. 

\subsection{Pre-training with GAPT}

In our second experiment, we retain the same GPT pre-training setup but incorporate the GAPT algorithm to test whether it offers a better solution to IBLM objective defined in Equation \ref{eq:iblm} , compared to a baseline model trained solely on the CE loss. MBE regularization during compression phase in GAPT is done from layer 2 to 9. 

\begin{figure}[h]
\centering

\begin{minipage}{0.48\textwidth}
    \centering
    \includegraphics[width=\linewidth]{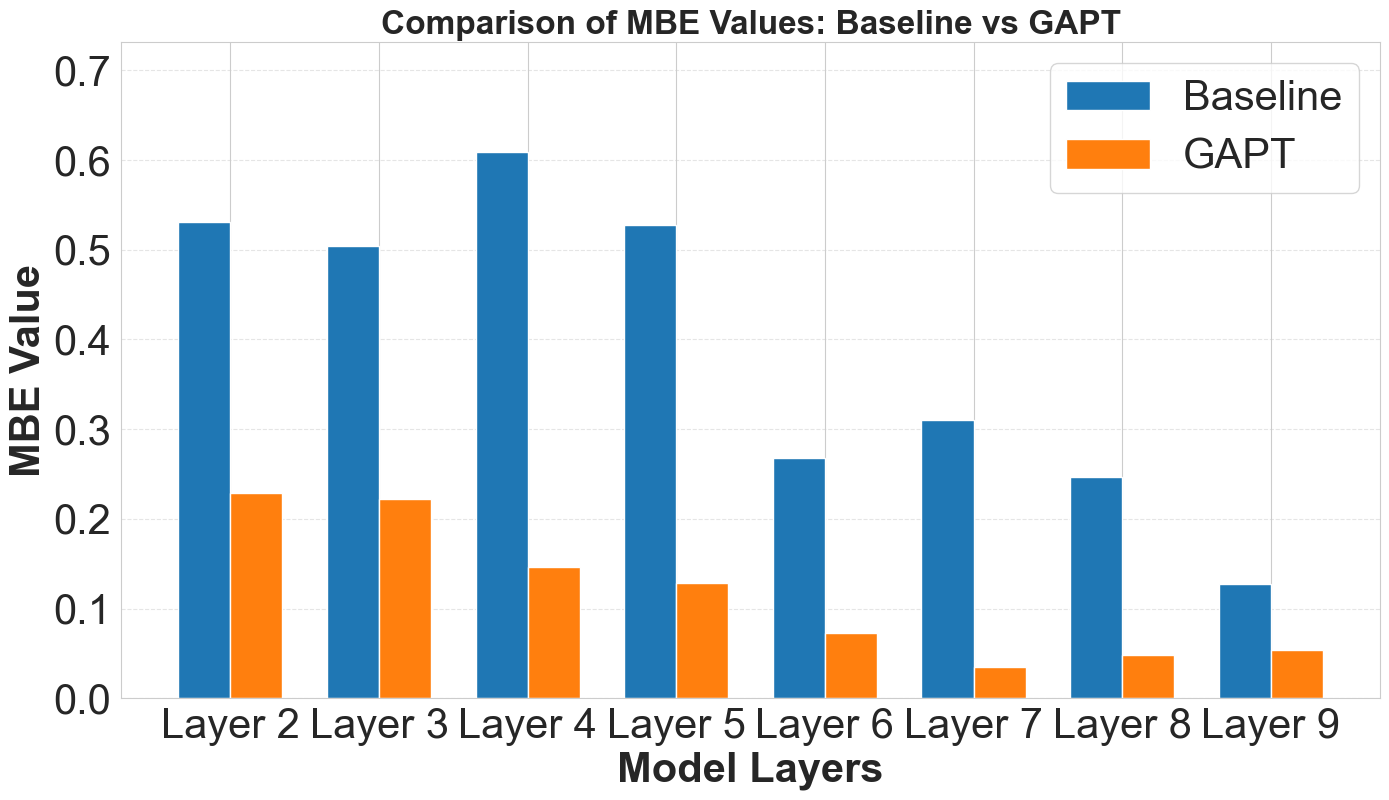}
\end{minipage}
\hfill
\begin{minipage}{0.48\textwidth}
    \centering
    \small
    \begin{tabular}{c|c|c|c}
        \toprule
        \textbf{Layer} & \textbf{Base} & \textbf{GAPT} & \textbf{Reduction} \\
        \midrule
        2 & 0.5313 & \textbf{0.2285} & \textcolor{green}{\scriptsize --56.99\%} \\
        3 & 0.5039 & \textbf{0.2217} & \textcolor{green}{\scriptsize --56.01\%} \\
        4 & 0.6094 & \textbf{0.1465} & \textcolor{green}{\scriptsize --75.96\%} \\
        5 & 0.5273 & \textbf{0.1279} & \textcolor{green}{\scriptsize --75.74\%} \\
        6 & 0.2676 & \textbf{0.0728} & \textcolor{green}{\scriptsize --72.81\%} \\
        7 & 0.3105 & \textbf{0.0344} & \textcolor{green}{\scriptsize --88.91\%} \\
        8 & 0.2471 & \textbf{0.0483} & \textcolor{green}{\scriptsize --80.44\%} \\
        9 & 0.1270 & \textbf{0.0542} & \textcolor{green}{\scriptsize --57.32\%} \\
        \midrule
        \rowcolor{blue!6}
        \textbf{Avg} & -- & -- & \textcolor{green}{\textbf{-70.52\%}} \\
        \bottomrule
    \end{tabular}
\end{minipage}

\vspace{0.5em}
\caption{Left: Layer-wise MBE for baseline vs. GAPT. Right: Per-layer MBE reduction with GAPT.}
\label{fig:gapt_pretrain_mbe}
\end{figure}

\begin{table}[h]
\centering
\setlength{\abovecaptionskip}{1em}  
\caption{Cross-entropy loss on the FineWeb validation set.}
\small

\begin{tabular}{l|c}
\toprule
\textbf{Model} & \textbf{CE Loss} \\
\midrule
Baseline     & 3.31 \\
\rowcolor{blue!6}
GAPT (Ours)  & \textbf{3.15} \textcolor{green}{\scriptsize (--4.8\%)} \\
\bottomrule
\end{tabular}

\label{tab:gapt_pretrain_ce}
\end{table}
As shown in Table \ref{tab:gapt_pretrain_ce}, GAPT reduces cross-entropy on the FineWeb validation set by $4.8\%$ compared to the CE-only baseline.  In addition to reducing test CE loss, GAPT also significantly compresses internal representations. Figure \ref{fig:gapt_pretrain_mbe} (left) shows the layer-wise MBE for both models. We observe consistent reductions across all regularized layers. The per-layer MBE values and their relative improvements are summarized in Figure \ref{fig:gapt_pretrain_mbe} (right). GAPT reduces MBE by an average of $70.5\%$ across layers 2–9 while improving validation cross-entropy by $4.8\%$. This suggests that explicitly alternating between memorization and compression phases offers an effective solution to the constrained optimization objective in IBLM (Equation \ref{eq:iblm}), and can exceed baseline training with cross-entropy target alone.

\subsection{Conflicting memory resolution}

Inspired by \cite{gonalez2020sleepprotectmemories}, which showed that sleep consolidation helps resolve memory conflicts, we design a synthetic experiment to test whether GAPT can mitigate representation interference between conflicting experiences. We use a 2-layer MLP \( f_\theta \) with randomly initialized parameters and define a symmetric shift \( \Delta\theta \). Inputs \( X_1, X_2 \in \mathbb{R}^{10} \) are sampled from Gaussians with shared variance but distinct means:

\begin{align*}
X_1 &\sim \mathcal{N}([1,1,1,1,1,0,0,0,0,0], \sigma^2 I), \\
X_2 &\sim \mathcal{N}([0,0,0,0,0,1,1,1,1,1], \sigma^2 I)
\end{align*}

The targets are defined as \( Y_1 = f_{\theta + \Delta\theta}(X_1) \) and \( Y_2 = f_{\theta - \Delta\theta}(X_2) \), producing two tasks with negatively aligned gradients. We compare GAPT against four baselines: single-task training, ordered training, mixed training, and GAPT applied to mixed batches with MBE regularization.

\begin{table}[h]
\centering
\setlength{\abovecaptionskip}{1em}  
\caption{Performance on conflicting experience learning. Lower L1/MBE and higher separation indicate better generalization and disentanglement.}
\small
\begin{tabular}{l|cc|cc|cc}
\toprule
\textbf{Strategy} & \textbf{L1 (Pos)} & \textbf{L1 (Neg)} & \textbf{MBE (Pos)} & \textbf{MBE (Neg)} & \textbf{Dist.} & \textbf{Sep. Ratio} \\
\midrule
Pos-only              & 0.02 & 0.71 & 0.18 & 0.45 & --   & --   \\
Neg-only              & 0.92 & 0.02 & 0.36 & 0.36 & --   & --   \\
Pos $\rightarrow$ Neg & 0.43 & 0.04 & 0.19 & 0.15 & 2.43 & 2.84 \\
Neg $\rightarrow$ Pos & 0.02 & 0.57 & 0.19 & 0.43 & 1.86 & 2.33 \\
Mixed                 & 0.03 & 0.03 & 0.10 & 0.22 & 3.66 & 4.11 \\
\rowcolor{blue!6}
GAPT + MBE (Ours)     & \textbf{0.03} & \textbf{0.03} & \textbf{0.02}\textcolor{green} {\scriptsize (--80\%)}& \textbf{0.02}\textcolor{green} {\scriptsize (--91\%)}& \textbf{6.64}\textcolor{green}{\scriptsize (+81\%)}& \textbf{8.08}\textcolor{green} {\scriptsize (+97\%)}\\
\bottomrule
\end{tabular}

\label{tab:conflict_summary}
\end{table}
Table \ref{tab:conflict_summary} summarizes results. Ordered learning, where the model is trained on one experience and then on the other, suffers from catastrophic forgetting: performance on the first task degrades significantly after exposure to the second. Mixed training alleviates this issue, achieving low L1 loss on both tasks and moderate representation separation. However, GAPT improves over mixed training in both respects: it maintains the same L1 accuracy while achieving a $97\%$ increase in separation ratio and a $91\%$ reduction in MBE. 

These results suggest that the compression encouraged by GAPT not only preserves generalization performance but also promotes the disentanglement of conflicting memories. Moreover, compression and separation emerge in tandem during training, closely resembling the consolidation behavior observed in biological neural systems during sleep. This supports the view that compression serves not only as a generalization mechanism, but also as a functional tool for resolving interference in memory \cite{gonalez2020sleepprotectmemories}.

\subsection{Arithmetic generalization}
To evaluate whether GAPT improves generalization in pre-training language models, we conduct a controlled experiment using a synthetic arithmetic dataset. We pre-train a GPT-2 model from scratch to perform integer multiplication. The training dataset contains 10 million multiplication equations between integers with 1–3 digits. For evaluation, we prepare two test sets: an in-domain (ID) set with 10,000 examples also from the 1–3 digit range, and an out-of-domain (OOD) set with 10,000 examples involving 4–6 digit multiplications. An additional OOD validation set with 1,000 examples is used for early stopping.

We tokenize the input and output sequences at the per-digit level and train the model for 1,750 iterations with a batch size of 16. Due to observed instability in OOD entropy, we adopt an early stopping strategy that halts training if validation loss increases by more than $20\%$ after iteration 800. 

\begin{figure}[t]
\centering
\small

\begin{minipage}[t]{\linewidth}
    \centering
    \includegraphics[width=0.75\linewidth]{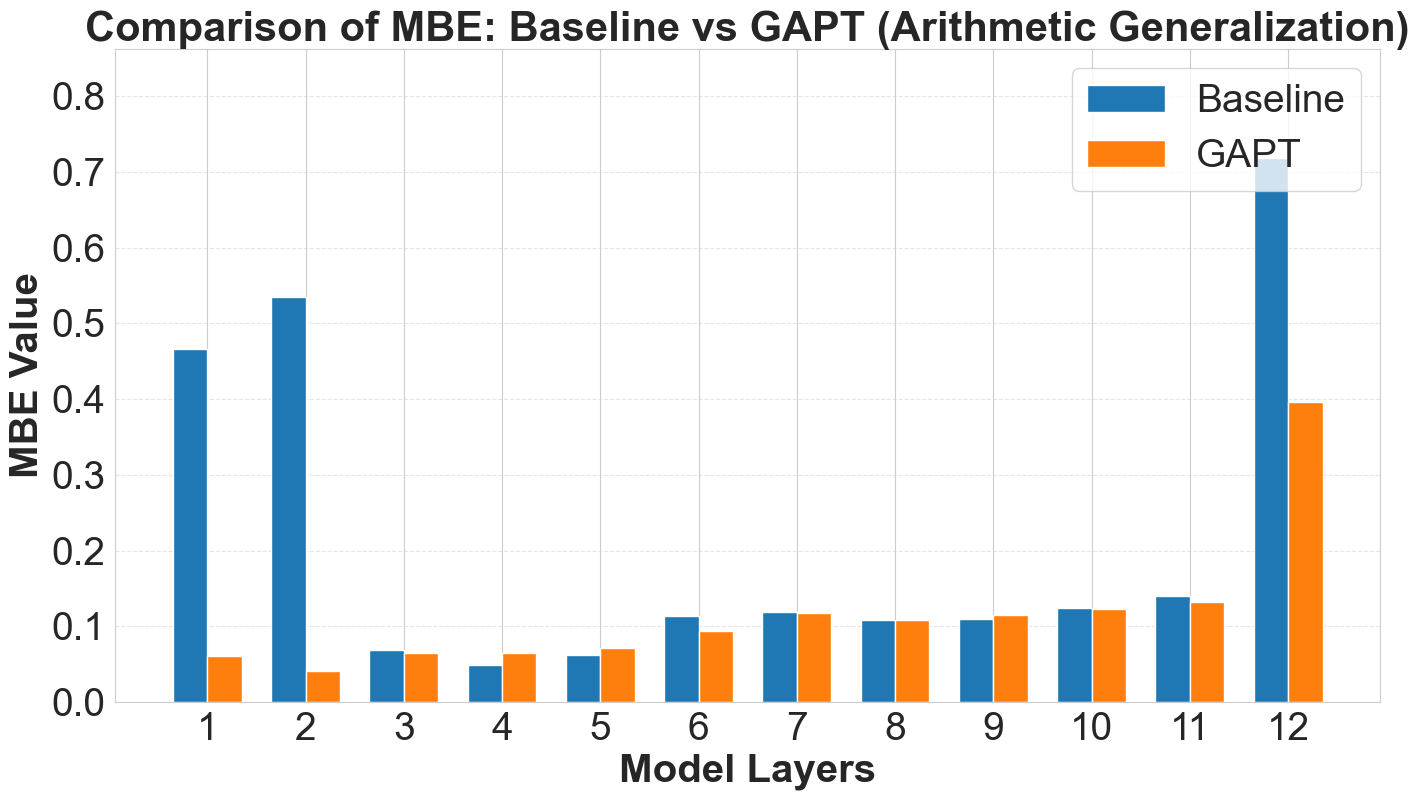}
    \vspace{-0.5em}
\end{minipage}

\vspace{1em}

\begin{minipage}[t]{0.75\linewidth}
    \centering
    \begin{tabular}{l|c|c|c}
        \toprule
        \textbf{Metric} & \textbf{Baseline (1085)} & \cellcolor{blue!6}\textbf{GAPT (Ours, 898)} & \textbf{Change} \\
        \midrule
        Entropy (ID)     & 0.010 & 0.011 & \textcolor{red}{\scriptsize +10\%} \\
        Entropy (OOD)    & 4.334 & \textbf{2.817} & \textcolor{green}{\scriptsize --35\%} \\
        Avg MBE (0–11)   & 0.218 & \textbf{0.115} & \textcolor{green}{\scriptsize --47\%} \\
        \bottomrule
    \end{tabular}
   
\end{minipage}

\vspace{0.5em}
\caption{Up: Comparison of baseline vs. GAPT on arithmetic generalization. Bottom: Arithmetic generalization performance summary. GAPT improves OOD generalization and yields more compact representations.}
\label{fig:gapt_arithmetic}
\end{figure}

As shown in Figure \ref{fig:gapt_arithmetic}, GAPT improves generalization substantially. It reduces OOD entropy by $35\%$ and average representation entropy (MBE) by $47\%$, while maintaining similar performance on the in-domain set. This supports our theoretical prediction that minimizing representation entropy leads to stronger generalization under the Information Bottleneck Language Modeling (IBLM) framework.

Interestingly, while MBE regularization is only applied to a subset of layers, we observe that GAPT achieves lower MBE even in unregularized layers (e.g., layers 0, 1, and 11), suggesting a degree of entropy compression generalization across the network. Notably, MBE in layer 1 is reduced by $92\%$, and in layer 11 by $45\%$.

\section{Relevant work}
\label{section:relevant_work}
The \textbf{Information Bottleneck (IB)} method was introduced in \cite{tishby2000information} to formalize the goal of retaining only task-relevant information in representations by minimizing entropy. A theoretical connection between generalization and the cardinality of discrete inputs was established in \cite{shamir2010boundongengap}, and later applied to deep networks in \cite{tishby2015deeplearninginformationbottleneck}. However, the entropy–generalization link remained incomplete due to the gap between cardinality and entropy measures. Empirical evidence of a two-phase learning dynamic—early memorization followed by entropy compression—was presented in \cite{shwartzziv2017openingblackboxdeep}. To quantify entropy in neural networks, matrix-based entropy (MBE) was proposed in \cite{giraldo2014entropy}, and later applied to LLMs in \cite{skean2025layerlayeruncoveringhidden}, where it correlated with embedding quality and revealed compression trends across checkpoints.

LLMs such as GPT-2 \cite{radford2019gpt2} generalize across tasks by scaling training data and parameters \cite{kaplan2020scalinglawsneurallanguage}, effectively compressing the training corpus \cite{huang2024compressionrepresentsintelligencelinearly, deletang2024languagemodelingcompression}. While post-training methods like instruction tuning \cite{ouyang2022traininglanguagemodelsfollow} improve usability, LLMs still struggle on out-of-distribution tasks such as reverse reasoning \cite{berglund2024reversalcurse, yu2024iterativegraphalignment} and multi-hop inference \cite{yu2024iterativegraphalignment}. Recent advancements in RL with verifiable rewards (RLVR) have improved mathematical and coding performance \cite{deepseekai2025deepseekr1}, though often by narrowing base model behavior \cite{yue2025doesreinforcementlearningreally}. Finally, curriculum-based tokenization growth has been combined with pre-training to improve compression \cite{yu2025scalingllmpretrainingvocabulary}.

\appendix

\section{Appendix}
\label{section:appendix}

\begin{theorem}[Minimum Probability Entropy Bound]
Let \( X \) be a discrete random variable with sample space \( \Omega \), where \( |\Omega| = n \). Suppose there exists a constant \( \alpha \in \left(0, \frac{1}{n}\right] \) such that \( P(X = x) \geq \alpha \) for all \( x \in \Omega \). Then the entropy of \( X \) is bounded below by:
\[
H(X) \geq -\left(1 - \alpha(n-1)\right)\log_2\left(1 - \alpha(n-1)\right) - (n-1)\alpha \log_2(\alpha).
\]
Furthermore, for sufficiently large \( n \) and small \( \alpha \) such that \(\beta =  \alpha n \ll 1 \), this bound approximates to:
\[
H(X) \geq  \beta \log_{2}(n)
\]
\label{thm:entropy-lower}
\end{theorem}

\begin{proof}
Under the constraint that \( P(X = x) \geq \alpha \) for all \( x \in \Omega \), the entropy
\[
H(X) = -\sum_{x \in \Omega} P(x)\log_2 P(x)
\]
is minimized when the distribution is as imbalanced as possible: one outcome has the highest allowed probability, and all others are assigned the minimum \( \alpha \). The worst-case distribution is:
\begin{align*}
P(x_1) &= 1 - \alpha(n - 1), \\
P(x_i) &= \alpha \quad \text{for } i = 2, \dots, n.
\end{align*}

The resulting entropy is:
\[
H(X) = -\left(1 - \alpha(n-1)\right)\log_2\left(1 - \alpha(n-1)\right) - (n-1)\alpha \log_2(\alpha).
\]

For small \( \alpha \) and large \( n \) such that \( \alpha n \ll 1 \), we approximate have:
\[
1 - \alpha(n-1) \approx 1 - \alpha n.
\]
Substituting this into the expression:
\begin{align*}
H(X) &\approx - (1 - \alpha n)\log_2(1 - \alpha n) - \alpha n \log_2(\alpha) \\
     &= - (1 - \alpha n)\log_2(1 - \alpha n) - \alpha n \log_{2}(\alpha n) + \alpha n \log_{2}(n) \\
     &= h(\alpha n) + \alpha n \log_2(n)
\end{align*}
where \( h(p) = -p\log_2 p - (1 - p)\log_2(1 - p) \) is the binary entropy function.

Since \( h(\alpha n) \geq 0 \) and \( \log_2(1/\alpha) \geq \log_2(n) \) for \( \alpha \leq 1/n \), we get:
\[
H(X) \geq \alpha n \log_2(n) = \beta \log_{2}(n)
\]
which completes the proof.
\end{proof}

\begin{repcor}[Entropy Lower Bound for Finite Discrete Random Variables]
Let \( X \) be a discrete random variable with finite support \( \Omega \), where \( |\Omega| = n \), and assume that \( P(X = x) > 0 \) for all \( x \in \Omega \). Then there exists a constant \( \beta \in (0,1] \) such that:
\[
H(X) \geq \beta \cdot \log_{2} |\Omega|.
\]
\end{repcor}

\begin{proof}
Let \( \varepsilon := \min_{x \in \Omega} P(X = x) \), which exists and is strictly positive since \( X \) is discrete with finite support. By the Minimum Probability Entropy Bound (Theorem~\ref{thm:entropy-lower}), we have:
\[
H(X) \geq -\left(1 - \varepsilon(n - 1)\right)\log_2\left(1 - \varepsilon(n - 1)\right) - (n - 1)\varepsilon \log_2(\varepsilon).
\]
For small \( \varepsilon \) and large \( n \), this approximates to:
\[
H(X) \geq \varepsilon n \log_2\left(\frac{1}{\varepsilon}\right),
\]
which is linear in \( n = |\Omega| \). Setting \( \beta := \varepsilon n \), the result follows:
\[
H(X) \geq \beta \cdot \log_{2} |\Omega|.
\]
\end{proof}

\section*{Broader Impact and Limitations}

\paragraph{Limitations.} While theoretically grounded, our proposed GAPT algorithm remains relatively simple. On large-scale pretraining tasks, it achieves less than a 5\% improvement over the baseline, falling short of fully realizing the theoretical benefits of compact representations for generalization as suggested by Theorem~\ref{thm:entropy-generalization}. Furthermore, scaling experiments are missing, which requires more computation resources but may realize more potential in this new dimension for improving generalization. Additionally, entropy minimization is only one approach to compression; other mechanisms—such as sparse activation—may offer stronger biological plausibility and efficiency. A further limitation is observed in the arithmetic generalization task, where GAPT exhibits instability in out-of-distribution performance across different training runs. There could be some data selection side piece missing for stabilizing generalization-oriented training runs. 

\paragraph{Broader Impact.} Enhancing generalization in AI systems holds the potential to enable models that can extrapolate, reason, and discover novel knowledge—key milestones toward artificial general intelligence. However, systems that generalize without explainability may pose greater risks than those that merely memorize. We caution that aggressive compression of internal representations, while beneficial for efficiency and abstraction, may also increase susceptibility to unintended behaviors such as spurious generalization or adversarial misuse.

\end{document}